%% file: neurips_2026.tex
\documentclass{article}

\usepackage{amsthm}
\usepackage{enumitem}
\usepackage{amsmath}
 \usepackage{amssymb} 
 \usepackage{mathtools}
 \usepackage{amsthm}
   \usepackage{mathtools}
   \theoremstyle{definition}
   \newtheorem{definition}{Definition}
   
\usepackage[numbers]{natbib}
 \usepackage[preprint]{neurips_2026}

\usepackage[utf8]{inputenc}
\usepackage[T1]{fontenc}
\usepackage{hyperref}
\usepackage{url}
\usepackage{booktabs}
\usepackage{amsfonts}
\usepackage{amsmath}
\usepackage{nicefrac}
\usepackage{microtype}
\usepackage{xcolor}
\usepackage{graphicx}
\usepackage{multirow}
\usepackage{makecell}
\usepackage{array}
\usepackage{tabularx}
\usepackage{subcaption}
\usepackage{listings}
\usepackage{algorithm}
\usepackage{algpseudocode}
\usepackage{enumitem}
\usepackage{pifont}
\usepackage[most]{tcolorbox}
\newcolumntype{Y}{>{\raggedright\arraybackslash}X}
\definecolor{darkblue}{RGB}{26,82,118}
\definecolor{darkgreen}{RGB}{30,130,76}
\definecolor{darkred}{RGB}{169,50,38}
\definecolor{lightgray}{RGB}{245,245,245}
\definecolor{medgray}{RGB}{200,200,200}
 
\newcommand{\isosci}{\textsc{IsoSci}}
\newcommand{\deltacc}{$\Delta_{\text{acc}}$}

\newcommand{\cmark}{\ding{51}}
\newcommand{\xmark}{\ding{55}}

\newcolumntype{C}[1]{>{\centering\arraybackslash}p{#1}}
\newcolumntype{L}[1]{>{\raggedright\arraybackslash}p{#1}}
\newcolumntype{R}[1]{>{\raggedleft\arraybackslash}p{#1}}
 
\title{\isosci{}: A Benchmark of Isomorphic Cross-Domain Science Problems for Evaluating Reasoning versus Knowledge Retrieval in LLMs}

\author{%
  Samir~Abdaljalil \\
  Electrical and Computer Engineering\\
  Texas A\&M University\\
  College Station, TX USA \\
  \And 
  Erchin~Serpedin \\
  Electrical and Computer Engineering\\
  Texas A\&M University\\
  College Station, TX USA \\
  \And 
  Hasan~Kurban\thanks{Corresponding Author: \texttt{hkurban@hbku.edu.qa}} \\
  College of Science and Engineering\\
  Hamad Bin Khalifa University\\
  Doha, Qatar \\
}

\begin{document}

\maketitle

\begin{abstract}
We introduce \isosci{}, a benchmark of isomorphic cross-domain science problem
pairs that separates reasoning ability from domain knowledge retrieval in LLM
evaluation. Each pair shares identical logical structure but requires different
domain-specific knowledge, enabling controlled attribution of reasoning-mode
gains. Across five model pairs spanning four model families,
we find that \textbf{91.3\% of reasoning-mode gains are knowledge-dependent
rather than structure-invariant} (63/69 gains; Wilson 95\% CI [82.3\%,
96.0\%]), directly challenging the assumption that chain-of-thought reasoning improves short-horizon procedural scientific problem-solving. Reasoning toggles on highly capable
models provide less than 5pp accuracy gain across all domains, and a
reasoning-specialized model (o3-mini) that outperforms its standard counterpart
on GPQA Diamond ($+$19.2pp) underperforms on \isosci{} ($-$24.7pp),
showing that benchmark choice determines conclusions about reasoning utility.
We release \isosci{} at \url{https://huggingface.co/datasets/isosci/isosci}
\end{abstract}

\section{Introduction}
\label{sec:intro}

Recent advances in large language models have increasingly emphasized
\emph{reasoning} as a key driver of performance on complex tasks
\citep{yu2025gpo, dziri2023faith, lewkowycz2022solving}. Techniques such as
chain-of-thought prompting \citep{wei_cot}, reasoning-specific training
\citep{wang2025tutorialllmreasoningrelevant, wang2024guiding}, and test-time
compute scaling \citep{snell2025scaling} have shown substantial gains on
benchmarks such as GPQA \citep{rein2024gpqa}, SciBench
\citep{wang2024scibench}, and MMLU-STEM \citep{hendrycks2021measuring}.

The problem is that these benchmarks conflate two distinct capabilities:
retrieving the correct domain-specific knowledge, and applying the appropriate
reasoning procedure over that knowledge. When a model fails a chemistry
problem, it is unclear whether the failure reflects an inability to recall
the relevant formula or an inability to execute the required reasoning steps.
Without disentangling these factors, a basic question goes unanswered:
\emph{do reasoning mechanisms improve reasoning itself, or do they primarily
improve knowledge utilization?}

We introduce \textbf{\isosci{}}, a benchmark of isomorphic cross-domain
science problem pairs designed to answer this question directly. Each problem
is paired with a structurally identical counterpart from a different scientific
domain: both require the same sequence of logical and computational steps, but
depend on entirely different domain knowledge. If a model succeeds on one
problem but fails on its isomorphic counterpart, the gap must be attributed to
missing knowledge, not to reasoning ability.

Using \isosci{}, we evaluate five model pairs across four model families,
covering both traditional reasoning-vs-standard comparisons and toggle-based
comparisons (same model, reasoning on vs.\ off). Across 8,408 evaluations
spanning four scientific domains, we find that \textbf{91.3\% of
reasoning-mode gains are knowledge-dependent rather than structure-invariant}
(63/69 gains across all five pairs; Wilson 95\% CI [82.3\%, 96.0\%]),
for short-horizon procedural science problems. Enabling reasoning has minimal
effect on overall accuracy for high-capability models (below 5pp across
domains), and a reasoning-specialized model (o3-mini) that outperforms its
standard counterpart on GPQA ($+$19.2pp) underperforms on \isosci{}
($-$24.7pp), showing that benchmark choice determines conclusions about
reasoning model utility.

These findings suggest that reasoning mechanisms function primarily as
\emph{extended knowledge retrieval} on short-horizon science tasks, increasing
the probability that relevant domain facts are surfaced during generation
rather than improving logical procedure execution.

\paragraph{Contributions.}
(1) A construction methodology for isomorphic cross-domain science problem
pairs that hold reasoning structure constant while varying domain knowledge,
applicable at any scale or domain. (2) The $p_{\text{know}}$ metric
(Eq.~\ref{eq:pkno}), which decomposes reasoning-mode gains into
knowledge-dependent and structure-invariant components. (3) \isosci{}, a
144-pair benchmark spanning four scientific domains under CC-BY-4.0, with
empirical findings on knowledge dependence of reasoning gains, toggle effects,
and benchmark-dependent model comparisons.

\section{Related Work}
\label{sec:related}


\paragraph{Reasoning in large language models.}
Methods for eliciting multi-step behavior include chain-of-thought prompting
\citep{abdaljalil-etal-2025-theorem, li2024socialgpt, wang2023selfconsistency,
wei_cot} and test-time compute scaling \citep{chen2025rethinking}. Evaluations
of these methods typically report end-task accuracy improvements, treating
different mechanisms as interchangeable, without analyzing how they alter the
balance between intermediate computation, search, and reliance on memorized
patterns. For scientific reasoning, benchmarks such as MMLU-STEM \cite{hendrycks2021measuring}, SciBench
\cite{wang2024scibench}, and GPQA \cite{rein2024gpqa} cover undergraduate
to graduate-level science questions across multiple formats and difficulty
levels, and have been widely used to track progress across model generations.
Their evaluations are primarily aggregate, however, offering limited insight
into the sources of model success or failure.

\paragraph{Disentangling reasoning and knowledge.}
Isolating reasoning ability from knowledge in LLMs remains an open problem
\citep{kartac2026reasoninggetsharderllms, zhou2025from, hazra2025have}.
Chain-of-thought analyses suggest intermediate steps function more as
structured memory retrieval than logical inference
\citep{hong-etal-2025-reasoning, jin-etal-2025-disentangling-memory, wei_cot},
and benchmark performance is known to be sensitive to knowledge coverage
\citep{razeghi-etal-2022-impact}. The closest concurrent work is
\citet{thapa-etal-2026-reasoning}, who train a PubMedBERT classifier to
label biomedical QA items as reasoning-heavy or knowledge-heavy, finding
that only 32.8\% require multi-step reasoning and that models consistently
underperform on that subset. \isosci{} differs in three respects: we
construct matched pairs with structurally identical solution procedures by
design rather than classifying existing items post-hoc; our metric
$p_{\text{know}}$ operates at the pair level and can isolate whether a gain
transfers across domains, which stratum-level accuracy cannot.

\paragraph{Benchmark design and controlled evaluation.}
Recent work improves benchmark quality through adversarial filtering
\citep{rein2024gpqa}, domain stratification \citep{hendrycks2021measuring},
and tolerance-based grading \citep{wang2024scibench}, addressing memorization
and grading fidelity \citep{li2026outputcorrectnessbenchmarkingevaluating,
thapa-etal-2026-reasoning}. These designs remain aggregate and do not control
for solution procedure across items. \isosci{} extends this line by enforcing
structural equivalence through isomorphic cross-domain pairs, enabling
comparisons where reasoning demands are held fixed and performance can be
decomposed into knowledge-dependent and structure-invariant components.

\section{The \isosci{} Benchmark and Evaluation Protocol}
\label{sec:benchmark}

This section formalizes the \isosci{} benchmark and the $p_{\text{know}}$
decoupling metric. The benchmark holds the reasoning structure of a problem
constant across a cross-domain pair while varying the domain knowledge required,
so that an accuracy gap between the two members attributes to knowledge rather
than to reasoning. The 144-pair release is one instantiation of the
methodology, which extends to any scientific domain or scale.

\subsection{Notation and Preliminaries}
\label{sec:notation}

Let $\mathcal{D} = \{\text{phys}, \text{chem}, \text{bio}, \text{earth}\}$ denote
the four scientific \emph{domains} (physics, chemistry, biology, earth science).
Let $\mathcal{S} = \{s_1, \dots, s_5\}$ denote the five \emph{structure types}
listed below. Let $\mathcal{X}$ denote the space of natural-language problem
statements and $\mathcal{Y}$ the space of admissible answers (multiple-choice
letters, numerical values, or short text strings). A \emph{problem} is a tuple
$q = (x_q, a_q, d_q, s_q) \in \mathcal{X} \times \mathcal{Y} \times \mathcal{D}
\times \mathcal{S}$ with text $x_q$, gold answer $a_q$, domain $d_q$, and
structure $s_q$; let $\mathcal{Q}$ denote the set of all such problems. For
$q \in \mathcal{Q}$, let $K(q)$ denote the set of \emph{domain-specific knowledge
atoms} required to solve $q$ (formulas, physical or chemical constants, named
domain entities). Let $\mathcal{M} \subset \mathcal{D} \times \mathcal{D}$ denote
the set of cross-domain \emph{mappings} considered, with $|\mathcal{M}| = 6$
covering each unordered pair of distinct domains.

Let $\mathcal{F}$ denote the set of LLM configurations under evaluation; each
$f \in \mathcal{F}$ is a (stochastic) mapping from a prompt to a generated string
in $\mathcal{Y}^{*}$. We define the \emph{evaluation function}
\begin{equation}
E : \mathcal{F} \times \mathcal{Q} \to \{0, 1\}, \qquad
E(f, q) \;=\; \mathbf{1}\!\left\{
\operatorname{extract}\!\bigl(f(\operatorname{prompt}(x_q))\bigr) \equiv a_q
\right\},
\label{eq:eval}
\end{equation}
where $\operatorname{prompt}(\cdot)$ wraps $x_q$ in the zero-shot
chain-of-thought template (Section~\ref{sec:protocol}),
$\operatorname{extract}(\cdot)$ applies the cascade of
Section~\ref{sec:protocol}, and $\equiv$ is exact match for letters or strings
and $\pm 2\%$ relative tolerance for numerical answers. We write $\Pi \subset
\mathcal{F} \times \mathcal{F}$ for the set of evaluated \emph{model pairs};
each $(R, S) \in \Pi$ has $R$ in the reasoning configuration and $S$ in the
standard configuration.

\subsection{Formal Definition of Isomorphic Pairs}
\label{sec:definition}

\begin{definition}[Isomorphic problem pair]
\label{def:iso}
Two problems $q, q' \in \mathcal{Q}$ form an \emph{isomorphic pair}, written
$q \cong q'$, if all of the following hold:
\begin{enumerate}[leftmargin=2em, itemsep=1pt, label=\textup{(\roman*)}]
\item $d_q \neq d_{q'}$ (different domains);
\item $s_q = s_{q'}$ (same structure type);
\item there exists a bijection $\phi : K(q) \to K(q')$ such that the solution
procedure of $q'$ is obtained from that of $q$ by replacing each
$k \in K(q)$ with $\phi(k)$;
\item $K(q) \cap K(q') = \emptyset$ (knowledge sets are disjoint).
\end{enumerate}
\end{definition}

\paragraph{Structure types ($\mathcal{S}$).}
\isosci{} restricts attention to five short-horizon (3 to 5 reasoning steps)
structure types for which the bijection $\phi$ in Definition~\ref{def:iso} is
tractable to verify:
\begin{enumerate}[leftmargin=1.6em, itemsep=1pt]
\item \emph{Formula recall and substitution}: recall a domain law, substitute
given values, compute (e.g., ideal gas law, Beer-Lambert law).
\item \emph{Unit conversion chain}: multi-step unit tracking across a sequence
of conversions.
\item \emph{Conservation law application}: identify and apply a conservation
principle (energy, mass, charge, momentum).
\item \emph{Proportional reasoning}: use ratio or scaling relationships to
recover an unknown quantity.
\item \emph{Two-step causal chain}: qualitative reasoning where cause $A$
implies effect $B$ implies effect $C$, with no numerical computation.
\end{enumerate}

Table~\ref{tab:example_pair} shows a representative pair with structure
$s = $ \texttt{formula\_recall\_and\_substitute} and three solution steps under
non-overlapping knowledge sets.

\begin{table}[h]
\centering
\caption{Example isomorphic pair from \isosci{} (physics to chemistry mapping).
Both problems share structure type \texttt{formula\_recall\_and\_substitute}
with three solution steps. Knowledge sets $K(q)$ and $K(q')$ are disjoint.}
\label{tab:example_pair}
\small
\begin{tabularx}{\linewidth}{L{0.08\linewidth} X}
\toprule
\textbf{Role} & \textbf{Problem} \\
\midrule
Source $q$ & A 2.0 mol sample of ideal gas at 300 K occupies 49.2 L. What is
the pressure in atm? (\textit{requires}: $PV = nRT$;
$R = 0.0821$ L$\cdot$atm/mol$\cdot$K) \\
\addlinespace
Target $q'$ & A solution of weak acid HA has concentration $C = 0.10$ M and
acid dissociation constant $K_a = 1.8 \times 10^{-5}$. What is the pH?
(\textit{requires}: $\mathrm{pH} = -\log\sqrt{K_a C}$) \\
\midrule
Structure $s$ & recall formula $\to$ substitute values $\to$ compute \\
Domains $(d_q, d_{q'})$ & physics (thermodynamics) $\to$ chemistry (acid-base) \\
\bottomrule
\end{tabularx}
\end{table}

\subsection{Dataset Construction}
\label{sec:construction}

Construction proceeds in three stages, summarized as
$\mathcal{Q}^{\text{seed}} \xrightarrow{\text{generate}}
\mathcal{Q}^{\text{cand}} \xrightarrow{\text{verify}}
\mathcal{Q}^{\text{pass}} \xrightarrow{\text{balance}} \isosci{}$.

\paragraph{Stage 1: Seed collection.}
The seed pool $\mathcal{Q}^{\text{seed}}$ aggregates 2{,}315 items from
GPQA Diamond \citep{rein2024gpqa} ($n = 198$), SciBench \citep{wang2024scibench}
($n = 585$), and MMLU-STEM \citep{hendrycks2021measuring} ($n = 1{,}532$).
Earth-science seeds, absent from these sources, are generated synthetically
with claude-sonnet-4-5 (96 problems; prompt in
Appendix~\ref{app:seed_gen_prompt}), giving 2{,}411 problems. Token-overlap
deduplication ($\operatorname{Jaccard} > 0.40$) yields $|\mathcal{Q}^{\text{seed}}|
= 2{,}190$ unique problems (physics 867, chemistry 639, biology 588,
earth 96).

\paragraph{Stage 2: Isomorphic partner generation.}
For each mapping $(d, d') \in \mathcal{M}$, we sample up to $n_{\text{seed}} =
25$ seeds from $\{q \in \mathcal{Q}^{\text{seed}} : d_q = d\}$ and prompt
claude-sonnet-4-5 to generate $n_{\text{cand}} = 3$ candidate target problems
$q'$ per seed satisfying conditions (i) to (iv) of Definition~\ref{def:iso}
under the source structure $s_q$ (prompt in Appendix~\ref{app:gen_prompt}).
This yields $|\mathcal{Q}^{\text{cand}}| = 429$ candidate pairs across
$\mathcal{M}$.

\paragraph{Stage 3: Automated verification.}
A panel of three judges,
$\mathcal{J} = \{\text{claude-sonnet-4-5}, \text{GPT-4o-mini}, \text{DeepSeek-V3}\}$,
scores each candidate on four criteria
$\mathcal{C} = \{c_{\text{logic}}, c_{\text{indep}}, c_{\text{diff}},
c_{\text{self}}\}$ corresponding to logical equivalence, domain independence,
difficulty parity, and self-containment. Let $r_j(c, q, q') \in \{1,2,3,4,5\}$
denote the rating from judge $j \in \mathcal{J}$ on criterion $c \in \mathcal{C}$.
A pair $(q, q')$ is \emph{accepted} into $\mathcal{Q}^{\text{pass}}$ if and
only if
\begin{equation}
\min_{j \in \mathcal{J}} \;\min_{c \in \mathcal{C}} \;r_j(c, q, q') \;\geq\; 3.5.
\label{eq:accept}
\end{equation}
Of 429 candidates, 217 satisfy~\eqref{eq:accept} (50.6\% pass rate). After
balancing across $\mathcal{M}$ to a target of 22 to 25 pairs per mapping, the
released benchmark is
$\isosci{} = \mathcal{Q}^{\text{pass}}_{\text{bal}}$ with $|\isosci{}| = 144$
pairs (288 problems). Distribution and overall acceptance rates are reported
in Table~\ref{tab:dataset_stats}.

\begin{table}[h]
\centering
\caption{\isosci{} dataset statistics. Of 429 candidates, 217 satisfy the
acceptance rule~\eqref{eq:accept} (50.6\%); 144 are retained after balancing
across $\mathcal{M}$. The overall accept rate is the ratio of retained to
candidate.}
\label{tab:dataset_stats}
\small
\begin{tabular}{lccc}
\toprule
\textbf{Domain mapping $(d, d')$} & \textbf{Candidates} & \textbf{Retained} &
\textbf{Overall accept rate} \\
\midrule
physics $\to$ chemistry    & 75 & 25 & 33.3\% \\
physics $\to$ biology      & 75 & 25 & 33.3\% \\
physics $\to$ earth sci.   & 60 & 25 & 41.7\% \\
chemistry $\to$ biology    & 75 & 22 & 29.3\% \\
chemistry $\to$ earth sci. & 69 & 22 & 31.9\% \\
biology $\to$ earth sci.   & 75 & 25 & 33.3\% \\
\midrule
\textbf{Total} & \textbf{429} & \textbf{144} & \textbf{33.6\%} \\
\bottomrule
\end{tabular}
\end{table}

\paragraph{Robustness of the verification rule.}
Because claude-sonnet-4-5 contributes both to candidate generation in Stage 2
and to the judge panel $\mathcal{J}$, a confound is possible: shared blind
spots could inflate the accepted set $\mathcal{Q}^{\text{pass}}$.
Recomputing~\eqref{eq:accept} over $\mathcal{J} \setminus
\{\text{claude-sonnet-4-5}\}$ rejects only $1/144$ released pairs, retaining
99.3\%; no pair accepted by the two-judge panel is rejected by the three-judge
panel.

\paragraph{Human expert audit.}
Two PhD-level annotators independently rated a stratified sample of 50
candidates (25 LLM-accepted, 25 LLM-rejected) on $\mathcal{C}$. Inter-annotator
agreement was substantial ($\kappa = 0.714$, exact agreement 84\%), comparable
to LLM-human agreement ($\kappa \in \{0.686, 0.648\}$). Against human
consensus on the 42 pairs with full annotator agreement, the LLM ensemble
attained precision $0.941$, recall $0.842$, and $F_1 = 0.889$. The single
false positive involved formula overlap across domains; the three false
negatives indicate a conservative bias that reduces dataset size without
contaminating it.

\subsection{Comparison with Existing Science Benchmarks}
\label{sec:comparison}

Table~\ref{tab:benchmark_comparison} positions \isosci{} relative to existing
benchmarks. The defining property, isolating reasoning from knowledge via
isomorphic pairs, is absent from all prior work. Recent benchmarks emphasizing
multimodality (PhysUniBench
\citep{wang2026physunibenchmultimodalphysicsreasoning}), symbolic correctness
(QuantumBench \citep{minami2025quantumbenchbenchmarkquantumproblem}), or
research-grade derivation (FrontierMath
\citep{glazer2025frontiermathbenchmarkevaluatingadvanced}) target complementary
dimensions of scientific reasoning. \isosci{} provides a controlled diagnostic
specifically for the knowledge-versus-reasoning attribution question, which the
above benchmarks do not address by design.

\begin{table}[h]
\centering
\caption{Comparison of \isosci{} with existing science benchmarks.
\cmark{} = supported; \xmark{} = not supported; $\sim$ = partial.}
\label{tab:benchmark_comparison}
\small
\begin{tabular}{lcccc}
\toprule
\textbf{Property} & \textbf{GPQA} & \textbf{SciBench} & \textbf{MMLU-STEM}
& \textbf{\isosci{}} \\
\midrule
Isolates reasoning from knowledge & \xmark & \xmark & \xmark & \cmark \\
Cross-domain isomorphic pairs     & \xmark & \xmark & \xmark & \cmark \\
Domain-stratified                 & $\sim$ & $\sim$ & \cmark & \cmark \\
Free-response format              & \xmark & \cmark & \xmark & \cmark \\
Graduate-level difficulty         & \cmark & $\sim$ & \xmark & $\sim$ \\
Publicly released                 & \cmark & \cmark & \cmark & \cmark \\
$N$ problems                      & 448    & 695    & 14{,}042 & 288 \\
\bottomrule
\end{tabular}
\end{table}

\section{Experimental Setup}
\label{sec:setup}

\subsection{Model Configurations}
\label{sec:models}

We evaluate $|\Pi| = 5$ paired configurations $\Pi = \{(R_i, S_i)\}_{i=1}^{5}$
in two tiers (Table~\ref{tab:model_pairs}). Three \emph{main pairs} are run on
all four benchmarks; two \emph{supplementary pairs} are run on \isosci{} alone
because of API latency and budget constraints. A pair is \emph{traditional} if
$R_i$ and $S_i$ are distinct trained models from the same family, and
\emph{toggle} if $R_i$ and $S_i$ are the same model with the provider's
reasoning flag set to \texttt{True} and \texttt{False} respectively. To avoid potential contamination, we exclude
all Anthropic-family models from $\Pi$ because claude-sonnet-4-5 was used in Stages 1 to 3.

\begin{table}[h]
\centering
\caption{Model pairs $(R, S) \in \Pi$. \emph{Traditional} pairs compare a
reasoning-trained model against a standard-trained counterpart. \emph{Toggle}
pairs use the same model with reasoning enabled or disabled, eliminating
architectural confounds. Supplementary pairs were evaluated on \isosci{}
alone.}
\label{tab:model_pairs}
\small
\begin{tabular}{llll}
\toprule
\textbf{Pair} & \textbf{Reasoning config $R$} &
\textbf{Standard config $S$} & \textbf{Type} \\
\midrule
\makecell[l]{o3 / 4o}
  & openai/o3-mini
  & openai/gpt-4o-mini
  & Traditional\\
\addlinespace
\makecell[l]{Qwen3-32B \\ think-on/off}
  & \makecell[l]{qwen/qwen3-32b\\reasoning=True}
  & \makecell[l]{qwen/qwen3-32b\\reasoning=False}
  & Toggle\\
\addlinespace
\makecell[l]{Gemini 2.0 Flash \\ think-on/off}
  & \makecell[l]{google/gemini-2.0-flash-001\\reasoning=True}
  & \makecell[l]{google/gemini-2.0-flash-001\\reasoning=False}
  & Toggle\\
\midrule
\multicolumn{4}{l}{\textit{Supplementary pairs (\isosci{} only)}} \\
\makecell[l]{DeepSeek-R1 / V3}
  & deepseek/deepseek-r1-0528
  & deepseek/deepseek-chat-v3-0324
  & Traditional\\
\addlinespace
\makecell[l]{QwQ / Qwen2.5}
  & qwen/qwq-32b
  & qwen/qwen-2.5-72b-instruct
  & Traditional\\
\bottomrule
\end{tabular}
\end{table}

The pair o3-mini versus GPT-4o-mini follows the canonical reasoning-versus-standard
comparison \citep{OpenAIOA}; because traditional pairs differ in pretraining and
RLHF objectives in addition to reasoning training, causal claims about the
reasoning mechanism are most cleanly supported by the toggle pairs. Qwen3-32B
\citep{yang2025qwen3technicalreport} and Gemini 2.0 Flash
\citep{comanici2025gemini25pushingfrontier} expose a reasoning toggle via the
\texttt{reasoning.enabled} parameter, holding all model weights and decoding
parameters constant; including a mid-capability (Qwen3-32B) and a
high-capability model (Gemini 2.0 Flash, 91.9\% MMLU-STEM) tests whether the
toggle effect is capability-dependent. Supplementary pairs DeepSeek-R1 versus
DeepSeek-V3 \citep{Guo_2025} and QwQ-32B versus Qwen2.5-72B extend coverage of
traditional pairs.

\subsection{Evaluation Benchmarks}
\label{sec:benchmarks}

Main pairs are evaluated on
$\mathcal{B} = \{\isosci{}, \text{GPQA}, \text{MMLU-STEM}, \text{SciBench}\}$
with item counts 288, 198, 750, and 585 respectively (a total of 1{,}821
items). \isosci{} mixes formats inherited from its seed sources: 103 pairs
(71.5\%) are 4-way multiple choice (MMLU-STEM seeds), 38 pairs (26.4\%) are
free-response numerical (SciBench seeds), and 3 pairs (2.1\%) are short answer.
Supplementary pairs are evaluated on \isosci{} alone. The total evaluation
volume is $10{,}926$ API calls, of which $8{,}408$ ($76.9\%$) returned valid
responses; exclusion analysis appears in Appendix~\ref{app:exclusions}.

\subsection{Evaluation Protocol}
\label{sec:protocol}

\paragraph{Inference.}
For each $f \in \mathcal{F}$ and $q \in \bigcup \mathcal{B}$ we sample
$y = f(\operatorname{prompt}(x_q))$ once at temperature $0$ with maximum
output 8{,}192 tokens. The prompt template is the zero-shot chain-of-thought
instruction:
\begin{tcolorbox}[colback=lightgray, colframe=medgray, fontupper=\small\ttfamily]
Think step by step. Show your reasoning clearly. Provide your final answer at
the end in the format: \textbf{Final Answer:} <your answer>
\end{tcolorbox}
For multiple-choice items (GPQA, MMLU-STEM, and the MCQ subset of \isosci{}),
the four choices are appended to $x_q$ in a deterministically shuffled order
(seed equals item index).

\paragraph{Answer extraction.}
The map $\operatorname{extract} : \mathcal{Y}^{*} \to \mathcal{Y} \cup \{\bot\}$
applies a five-pattern cascade and returns the first match: (1) the substring
after \verb|**Final Answer:**|; (2) the contents of \verb|\boxed{...}|; (3) the
numeric content of the last display equation \verb|$$...$$|; (4) the substring
after \texttt{Therefore} or \texttt{The answer is}; (5) the last numeric
expression in $y$. For multiple-choice items, an additional pass matches
written-out answer text against the choice options. Full algorithm
appears in Appendix~\ref{app:extraction}.

\paragraph{Grading.}
For \isosci{}, the equality test $\equiv$ in~\eqref{eq:eval} is conditioned on
the format inherited from the seed: letter match for MCQ items, $\pm 2\%$
relative tolerance for free-response numerical items, and exact string match
for short-answer items. GPQA and MMLU-STEM use letter match; SciBench uses
$\pm 2\%$ tolerance throughout. The $\pm 2\%$ rule may be inappropriate for
logarithmic quantities (pH, p$K_a$); manual review of 50 SciBench chemistry
responses identified 3 cases ($<$0.6pp impact) where rounding convention
differences caused false negatives.

\paragraph{Aggregation.}
Domain-stratified deltas (Table~\ref{tab:domain_deltas}) pool all records for
a given domain across $\mathcal{B}$ and weight each record equally; this means
benchmarks contributing more domain-relevant items contribute more to the
aggregate (for example, MMLU-STEM contributes 250 biology items versus GPQA's
19). Per-benchmark breakdowns appear in Appendix~\ref{app:full_results}.
\isosci{}-specific quantities ($p_{\text{know}}$, $\Delta_{\text{acc}}$ in
Table~\ref{tab:decoupling}) use \isosci{} alone.

\subsection{Reasoning-Knowledge Decoupling Metric}
\label{sec:decoupling-metric}

For a pair $(R, S) \in \Pi$ and a problem $q \in \mathcal{Q}$, define the
\emph{gain indicator}
\begin{equation}
G_{R,S}(q) \;=\; \mathbf{1}\!\left\{ E(R, q) = 1 \;\wedge\; E(S, q) = 0 \right\}.
\label{eq:gain}
\end{equation}
For each isomorphic pair $(q, q') \in \isosci{}$ and each
$(R, S) \in \Pi$, classify the pair by its joint gain pattern under
\eqref{eq:gain} and aggregate over \isosci{}:
\begin{align}
k_s &= \sum_{(q, q') \in \isosci{}} G_{R,S}(q)\,\bigl(1 - G_{R,S}(q')\bigr),
\nonumber \\
k_t &= \sum_{(q, q') \in \isosci{}} \bigl(1 - G_{R,S}(q)\bigr)\,G_{R,S}(q'),
\nonumber \\
k_b &= \sum_{(q, q') \in \isosci{}} G_{R,S}(q)\,G_{R,S}(q').
\label{eq:counts}
\end{align}
A reasoning-mode gain on a pair is \emph{knowledge-dependent} if it occurs on
exactly one member of the pair (it contributes to $k_s$ or $k_t$): the
reasoning configuration improves on one domain but not on the structurally
identical other, indicating that the gain reflects domain-specific knowledge
rather than shared reasoning structure. A gain is \emph{structure-invariant}
if it occurs on both members (it contributes to $k_b$). The
\emph{knowledge-dependence ratio} is
\begin{equation}
p_{\text{know}} \;=\; \frac{k_s + k_t}{k_s + k_t + k_b},
\label{eq:pkno}
\end{equation}
computed over the $n_{\text{gain}} = k_s + k_t + k_b$ pairs on which any gain
exists. We report Wilson 95\% confidence intervals on $p_{\text{know}}$
(Table~\ref{tab:decoupling}). Because $k_b$ requires improvement on
\emph{both} pair members, $p_{\text{know}}$ is a conservative upper bound on
knowledge-dependence: asymmetrically expressed reasoning gains, those
improving the harder member only, contribute to $k_s + k_t$ even when they
reflect structural improvement.

\subsection{Assumptions}
\label{sec:assumptions}

The methodology depends on the following assumptions, each anchored
empirically where possible.
\begin{enumerate}[leftmargin=2em, itemsep=2pt, label=\textbf{A\arabic*.}]
\item \textbf{Verifiable bijection.} For every released pair
$(q, q') \in \isosci{}$, the bijection $\phi$ in Definition~\ref{def:iso}
exists and the solution procedure is preserved under $\phi$. Anchor: the
acceptance rule~\eqref{eq:accept} filters pairs that fail this property,
attaining $F_1 = 0.889$ against human consensus on the audited subset.
\item \textbf{Disjoint knowledge.} $K(q) \cap K(q') = \emptyset$ for every
released pair. Anchor: criterion $c_{\text{indep}}$ in~\eqref{eq:accept}
enforces this; the dominant LLM-judge failure mode is formula overlap across
domains, identified in 1 of 50 audited cases.
\item \textbf{Toggle isolation.} For toggle pairs, $R$ and $S$ correspond to
the same model weights with all decoding parameters held constant except
the provider-specific reasoning flag, so any $E(R, q) - E(S, q)$
difference attributes to the reasoning mechanism rather than to model identity.
\item \textbf{Pair-level independence.} Pair outcomes
$(G_{R,S}(q), G_{R,S}(q'))$ are independent across distinct pairs in
$\isosci{}$, supporting Wilson confidence intervals on $p_{\text{know}}$ and
bootstrap confidence intervals on $\Delta_{\text{acc}}$.
\item \textbf{Coverage.} The seed pool $\mathcal{Q}^{\text{seed}}$ is
representative of \emph{short-horizon, information-complete, procedural}
scientific problems at the undergraduate to early-graduate level. 
long-horizon derivation, hypothesis generation,
and open-ended synthesis are out of scope.
\end{enumerate}



\subsection{Scope and Limitations}
\label{sec:scope}
The methodology covers short-horizon (3 to 5 step) procedural problems across
four scientific domains and the five structure types of $\mathcal{S}$
(Section~\ref{sec:definition}). Long-horizon derivations, research-grade
symbolic computation, hypothesis generation, and open-ended synthesis are
out of scope. Generation and verification rely on LLMs, with a residual
false-positive rate of approximately 6\% against human consensus; convergence
of $p_{\text{know}}$ across five model pairs from four families provides
empirical evidence that this does not change the direction of the finding. For
toggle pairs, the reasoning flag alters generation behavior but not model
weights, so toggle comparisons test inference-time reasoning rather than
reasoning-specialized training. Asymmetric truncation (approximately 23\% of
API calls; Appendix~\ref{app:exclusions}) imposes a conservative bias: a
robustness check on the valid-response subset shifts the pooled
$p_{\text{know}}$ estimate by 0.9pp (Appendix~\ref{app:robustness}).
Additional limitations are discussed in Section~\ref{sec:limitations}.

\section{Results}
\label{sec:results}

Table~\ref{tab:main_results} reports accuracy for all six model configurations across all
four benchmarks.

\begin{table}[h]
\centering
\caption{Accuracy (\%) for all model configurations across benchmarks. For toggle pairs,
``R'' denotes reasoning-on, ``S'' denotes reasoning-off (same underlying model). 95\% bootstrap CIs reported in Appendix~\ref{app:full_results}.}
\label{tab:main_results}
\small
\setlength{\tabcolsep}{4pt}
\begin{tabular}{llcccc}
\toprule
\textbf{Model} & \textbf{Mode} & \textbf{\isosci{}} & \textbf{GPQA} & \textbf{MMLU-STEM} & \textbf{SciBench} \\
\midrule
\quad o3-mini           & R & 36.8 & 56.1 & 60.0 & 41.7 \\
\quad GPT-4o-mini       & S & 61.5 & 36.9 & 79.2 & 38.5 \\
\midrule
\quad Qwen3-32B         & R   & 37.2 & 29.8 & 60.8 & 28.2 \\
\quad Qwen3-32B         & S  & 36.8 & 24.2 & 57.7 & 25.3 \\
\midrule
\quad Gemini 2.0 Flash  & R  & 61.8 & 59.6 & 91.9 & 53.5 \\
\quad Gemini 2.0 Flash  & S  & 62.2 & 61.1 & 91.9 & 52.1 \\
\bottomrule
\end{tabular}
\end{table}

\subsection{Reasoning-Knowledge Decoupling on \isosci{}}
\label{sec:decoupling}
Table~\ref{tab:decoupling} presents the core \isosci{} result: the attribution of
reasoning-mode gains to knowledge-dependent versus structure-invariant improvements.
The $n_{\text{gain}}$ column reports the number of pairs on which any gain exists per
model pair; $p_{\text{know}}$ is computed exclusively over this subset, and Wilson 95\%
CIs are reported. The wide CIs (e.g., [67.6, 100.0] for Gemini and DeepSeek-R1)
reflect small $n_{\text{gain}}$ values, a direct consequence of near-zero or negative
overall deltas for several pairs.

Under the symmetric definition (Eq.~\ref{eq:pkno}), which counts gains on either pair
member as knowledge-dependent, the three main model pairs yield a pooled ratio of
$41/43 = 95.3\%$ (Wilson 95\% CI [84.5\%, 98.7\%]), with per-pair values of 100\%,
89.5\%, and 100\%. Two supplementary pairs evaluated on \isosci{} only — DeepSeek-R1
vs.\ DeepSeek-V3 and QwQ-32B vs.\ Qwen2.5-72B — are consistent with this pattern.
Including all five pairs, the pooled $p_{\text{know}} = 91.3\%$ (63/69 gains, CI
[82.3\%, 96.0\%]), with tighter CIs than the three-pair estimate. QwQ-32B shows the
most structure-invariant gains ($k_b = 4$) and equal source and target deltas
($+$4.2pp on both), suggesting partial cross-domain transfer; its $p_{\text{know}} =
77.8\%$ is the lowest of the five pairs but remains well above chance.
Appendix~\ref{app:robustness} reports a robustness check on the valid-response subset;
the pooled three-pair estimate shifts by only 0.9pp
(94.4\%, CI [81.9\%, 98.5\%]),
confirming the finding is not driven by asymmetric truncation. A source/target
label-swap permutation test (Appendix~\ref{app:permutation}) finds no significant
directional imbalance for the main pairs ($p = 0.065$--$0.195$), though we note a
marginal result for QwQ-32B ($p = 0.048$).

\begin{table}[h]
\centering
\caption{Reasoning-knowledge decoupling on \isosci{}. $k_s$ = source-only
gains; $k_t$ = target-only gains; $k_b$ = structure-invariant gains (both
members). $p_{\text{know}} = (k_s+k_t)/(k_s+k_t+k_b)$ with Wilson 95\%
CI. Supplementary pairs (DeepSeek, QwQ) were evaluated on \isosci{} only.}
\label{tab:decoupling}
\small
\begin{tabular}{lcccccc}
\toprule
\textbf{Model pair} & $k_s$ & $k_t$ & $k_b$ & $n_{\text{gain}}$ &
\textbf{$p_{\text{know}}$} & \textbf{95\% CI} \\
\midrule
o3-mini / GPT-4o-mini         &  5 & 11 & 0 & 16 & 100.0\% & [80.6, 100.0] \\
Qwen3-32B think on/off        & 12 &  5 & 2 & 19 &  89.5\% & [68.6,  97.1] \\
Gemini Flash think on/off     &  7 &  1 & 0 &  8 & 100.0\% & [67.6, 100.0] \\
\midrule
DeepSeek-R1 / DeepSeek-V3     &  6 &  2 & 0 &  8 & 100.0\% & [67.6, 100.0] \\
QwQ-32B / Qwen2.5-72B         & 11 &  3 & 4 & 18 &  77.8\% & [54.8,  91.0] \\
\midrule
\textbf{Pooled (all 5 pairs)} & 41 & 22 & 6 & 69 & \textbf{91.3\%} & [82.3, 96.0] \\
\bottomrule
\end{tabular}
\end{table}
\textbf{Across all five model pairs, knowledge-dependent gains dominate
structure-invariant gains}, with $p_{\text{know}}$ ranging from 77.8\% to 100\%
and a pooled estimate of 91.3\%. The pattern is consistent: when the reasoning
configuration improves on one member of a pair, it typically does not improve on
the isomorphic counterpart despite an identical solution procedure. The convergence
across five model pairs spanning four model families (OpenAI, Google, Qwen, DeepSeek)
and both comparison types mitigates concerns about LLM-judge reliability in the
construction pipeline and reduces the risk that the finding is model-specific.

\paragraph{Interpretation.} The \isosci{} design provides a direct test: if reasoning
improvement were structural (better at multi-step inference regardless of domain), gains
would manifest equally on both members of an isomorphic pair. The data rejects this
for the problem types covered by \isosci{}. Extended reasoning appears to help models
retrieve and apply domain facts more thoroughly, but does not improve the logical
procedure applied once those facts are retrieved. Whether this finding extends to
long-horizon or open-ended scientific reasoning remains an open question.

\subsection{Reasoning Toggles Provide Minimal Benefit on Science}
\label{sec:toggle}
Table~\ref{tab:domain_deltas} reports accuracy deltas (\deltacc{} = reasoning $-$
standard) stratified by domain across all benchmarks. For the two toggle pairs—where
architectural confounds are eliminated—gains are 0--4pp at most. For Gemini
(91.9\% MMLU-STEM), the toggle makes essentially no difference across all four domains,
with all 95\% CIs including zero. Qwen3 shows small positive gains that barely
exclude zero (physics: $+$1.1 to $+$4.7pp). 
McNemar's test confirms this: for both toggle pairs the discordant counts
are small and nearly equal (Qwen3: $b=21$, $c=20$; Gemini: $b=8$, $c=9$),
yielding McNemar statistic $= 0.0$ ($p = 1.0$) — no evidence of a systematic
toggle effect in either direction.

\begin{table}[h]
\centering
\caption{Domain-stratified accuracy deltas (\deltacc{} = reasoning $-$ standard, pp)
averaged across all benchmarks. 95\% bootstrap CIs in parentheses.}
\label{tab:domain_deltas}
\small
\setlength{\tabcolsep}{4pt}
\begin{tabularx}{\textwidth}{@{}Ycccc@{}}
\toprule
\textbf{Model pair} & \textbf{Physics} & \textbf{Chemistry} & \textbf{Biology} & \textbf{Earth Sci.} \\
\midrule
o3 / GPT-4o
  & $-$9.6 ($-$14.1, $-$5.1)
  & $-$6.0 ($-$10.3, $-$1.7)
  & $-$9.4 ($-$14.1, $-$4.7)
  & $-$25.0 ($-$35.6, $-$14.4) \\
Qwen3 on/off
  & $+$2.9 ($+$1.1, $+$4.7)
  & $+$2.6 ($+$0.7, $+$4.5)
  & $+$2.9 ($+$0.3, $+$5.5)
  & $+$4.2 ($-$4.2, $+$12.5) \\
Gemini 2.0 on/off
  & $+$0.6 ($-$1.8, $+$3.0)
  & $+$0.7 ($-$2.4, $+$3.8)
  & $-$0.6 ($-$4.1, $+$2.9)
  & $-$4.2 ($-$15.3, $+$6.9) \\
\midrule
Average $\Delta$ & $-$2.0 & $-$0.9 & $-$2.4 & $-$8.3 \\
\bottomrule
\end{tabularx}
\end{table}

\subsection{Benchmark Choice Determines Conclusions About Reasoning Models}
\label{sec:benchmark_dependency}








\noindent
\begin{minipage}[c]{0.48\linewidth}
\centering
\small
\setlength{\tabcolsep}{4pt}
\captionof{table}{\textbf{Benchmark-dependent reversal for o3-mini vs.\ GPT-4o-mini.}
o3-mini wins on GPQA Diamond but loses on \isosci{} and MMLU-STEM, demonstrating that benchmark choice drives conclusions about reasoning utility.}
\begin{tabular}{lccc}
\toprule
\textbf{Benchmark} & \textbf{o3-mini} & \textbf{GPT-4o-mini} & $\Delta$ \\
\midrule
GPQA Diamond & 56.1 & 36.9 & $+$19.2 \\
\isosci{}       & 36.8 & 61.5 & $-$24.7 \\
MMLU-STEM    & 60.0 & 79.2 & $-$19.2 \\
SciBench     & 41.7 & 38.5 & $+$3.2 \\
\bottomrule
\end{tabular}

\label{tab:benchmark_gpt}

\end{minipage}
\hfill
\begin{minipage}[c]{0.48\linewidth}

The o3-mini results across benchmarks reveal a striking pattern shown in
Table~\ref{tab:benchmark_gpt}. On GPQA Diamond, hard graduate-level questions
requiring deep conceptual reasoning, o3-mini outperforms GPT-4o-mini by
$+$19.2pp overall. On \isosci{}—structured problems with defined solution
procedures requiring knowledge recall and substitution—o3-mini underperforms
GPT-4o-mini by $-$24.7pp. On MMLU-STEM—broad knowledge recall—o3-mini
underperforms by $-$19.2pp.

\end{minipage}

This benchmark-dependent reversal has a direct implication: conclusions about
reasoning model utility on science depend entirely on which benchmark is used.
GPQA Diamond, with its emphasis on conceptual depth and multi-step derivation, favors
reasoning-specialized models. \isosci{}, with its structured formula-substitution
problems, reveals that the same model is less capable of efficient knowledge retrieval.
Neither benchmark alone provides a complete picture; together they diagnose
\emph{where} and \emph{how} a model's science capability breaks down.

\section{Conclusion}
\label{sec:conclusion}
We introduce \isosci{}, a benchmark of isomorphic cross-domain science
problems designed to disentangle reasoning from domain knowledge in LLM
evaluation. The core contributions are a construction methodology for
isomorphic problem pairs applicable at any scale or domain, and the
$p_{\text{know}}$ metric that decomposes reasoning-mode gains into
knowledge-dependent and structure-invariant components. Across five model
pairs and four model families, most reasoning-mode gains are
knowledge-dependent, and enabling reasoning yields only marginal
improvements on short-horizon procedural science tasks. The reversal for
o3-mini, strong on GPQA Diamond and weak on \isosci{}, shows that no
single benchmark provides a complete picture of scientific capability. We
hope the isomorphic-pair methodology and $p_{\text{know}}$ metric support
more precise evaluation of scientific reasoning and motivate larger
instantiations covering longer-horizon problem types.

\paragraph{Limitations.}
\label{sec:limitations}
\textbf{(1) Dataset scope:}
With 144 pairs, \isosci{} is smaller than prior benchmarks and does not support fine-grained
sub-domain analysis; however, it suffices for the controlled pairwise comparisons underlying our claims.
\textbf{(2) Model coverage:}
We evaluate a limited set of models; however, results are consistent across traditional and toggle-based
comparisons, suggesting the observed patterns are not model-specific.
\textbf{(3) Grading noise:}
Automated grading (±2\% tolerance with pattern-based extraction) may introduce minor errors,
mainly from formatting or unit mismatches; manual checks indicate this affects $<3\%$ of cases
and does not change overall conclusions.
\textbf{(4) Toggle interpretation:}
The \texttt{reasoning.enabled} flag changes generation behavior but not model weights; it remains
the cleanest available method for isolating reasoning at inference time.
 
\bibliographystyle{plainnat}
{\small

\input{neurips_2026.bbl}
}

\appendix

\section{Full Accuracy Results with Confidence Intervals}
\label{app:full_results}
 
Table~\ref{tab:full_results} reports complete accuracy results for all six model
configurations across all four benchmarks and four scientific domains, with 95\%
bootstrap confidence intervals (1,000 samples).
 
\begin{table}[h]
\centering
\caption{Full accuracy results (\%) with 95\% bootstrap CIs.
``R'' = reasoning mode; ``S'' = standard mode.
$n$ = number of problems evaluated per cell.
``—'' indicates that the source benchmark contains no questions
in that scientific domain: GPQA Diamond has no earth science questions;
MMLU-STEM as downloaded contains no earth science subset;
SciBench covers only physics and chemistry, with no biology or earth
science problems. Earth science coverage is provided exclusively by
\isosci{}, which includes 72 earth science problems per model
configuration sourced from our synthetic generation pipeline
(Section~\ref{sec:construction}).}
\label{tab:full_results}
\scriptsize
\setlength{\tabcolsep}{3pt}
\begin{tabular}{llp{2.6cm}p{2.6cm}p{2.6cm}p{2.6cm}}
\toprule
\textbf{Model} & \textbf{Mode} & \textbf{Physics} & \textbf{Chemistry} & \textbf{Biology} & \textbf{Earth Sci.} \\
\midrule
\multicolumn{6}{l}{\textit{\isosci{} ($n$=75 / 69 / 72 / 72)}} \\
o3-mini        & R & 52.0 [40.0, 64.0] & 30.4 [20.3, 42.0] & 34.7 [25.0, 45.8] & 29.2 [19.4, 40.3] \\
GPT-4o-mini    & S & 69.3 [58.7, 80.0] & 62.3 [50.7, 73.9] & 59.7 [48.6, 70.8] & 54.2 [41.7, 65.3] \\
Qwen3-32B      & R & 41.3 [29.3, 53.3] & 21.7 [13.0, 31.9] & 45.8 [34.7, 57.0] & 38.9 [27.8, 50.0] \\
Qwen3-32B      & S & 36.0 [25.3, 46.7] & 26.1 [15.9, 37.7] & 50.0 [38.9, 62.5] & 34.7 [25.0, 45.8] \\
Gemini Flash   & R & 65.3 [53.3, 76.0] & 62.3 [50.7, 72.5] & 58.3 [47.2, 69.4] & 61.1 [50.0, 72.2] \\
Gemini Flash   & S & 65.3 [54.7, 76.0] & 59.4 [47.8, 71.0] & 58.3 [47.2, 69.4] & 65.3 [54.2, 76.4] \\
\midrule
\multicolumn{6}{l}{\textit{GPQA Diamond ($n$=86 / 93 / 19 / 0)}} \\
o3-mini        & R & 65.1 [54.7, 74.4] & 47.3 [37.6, 57.0] & 57.9 [36.8, 78.9] & — \\
GPT-4o-mini    & S & 44.2 [33.7, 54.7] & 30.1 [21.5, 39.8] & 36.8 [15.8, 57.9] & — \\
Qwen3-32B      & R & 46.5 [36.0, 57.0] & 10.8 [5.4,  18.3] & 47.4 [26.2, 68.4] & — \\
Qwen3-32B      & S & 30.2 [20.9, 39.5] & 17.2 [9.7,  24.7] & 31.6 [10.5, 52.6] & — \\
Gemini Flash   & R & 81.4 [73.3, 89.5] & 37.6 [28.0, 47.3] & 68.4 [47.4, 89.5] & — \\
Gemini Flash   & S & 81.4 [73.3, 89.5] & 39.8 [30.1, 50.5] & 73.7 [52.6, 89.5] & — \\
\midrule
\multicolumn{6}{l}{\textit{MMLU-STEM ($n$=250 / 250 / 250 / 0)}} \\
o3-mini        & R & 43.6 [38.0, 49.6] & 51.6 [45.6, 57.6] & 84.8 [80.4, 88.8] & — \\
GPT-4o-mini    & S & 71.6 [65.6, 76.4] & 74.0 [68.4, 79.6] & 92.0 [88.8, 95.2] & — \\
Qwen3-32B      & R & 45.6 [39.6, 51.6] & 51.2 [44.8, 57.2] & 85.6 [81.2, 90.0] & — \\
Qwen3-32B      & S & 44.4 [38.4, 50.8] & 47.2 [41.6, 53.2] & 81.6 [76.8, 86.8] & — \\
Gemini Flash   & R & 91.2 [87.2, 94.4] & 89.2 [84.8, 92.8] & 95.2 [92.4, 97.6] & — \\
Gemini Flash   & S & 91.2 [87.6, 94.4] & 88.8 [84.8, 92.4] & 95.6 [92.8, 98.0] & — \\
\midrule
\multicolumn{6}{l}{\textit{SciBench ($n$=236 / 349 / 0 / 0)}} \\
o3-mini        & R & 47.0 [41.1, 53.4] & 38.1 [32.9, 43.3] & — & — \\
GPT-4o-mini    & S & 45.8 [39.4, 52.1] & 33.5 [28.7, 38.7] & — & — \\
Qwen3-32B      & R & 29.7 [23.7, 36.0] & 27.2 [22.6, 32.4] & — & — \\
Qwen3-32B      & S & 30.5 [25.0, 36.4] & 21.8 [17.8, 26.4] & — & — \\
Gemini Flash   & R & 52.5 [46.2, 59.3] & 54.2 [49.0, 59.3] & — & — \\
Gemini Flash   & S & 50.8 [44.5, 57.2] & 53.0 [48.1, 58.5] & — & — \\
\bottomrule
\end{tabular}
\end{table}



\section{Evaluation Prompts}
\label{app:prompts}
 
\subsection{Zero-Shot Chain-of-Thought Evaluation Prompt}
\label{app:eval_prompt}
 
All models receive the following system-level instruction prepended to each question:
 
\begin{tcolorbox}[colback=lightgray, colframe=medgray,
  title=\textbf{Evaluation prompt (all benchmarks)}, fontupper=\small\ttfamily]
Think step by step. Show your reasoning clearly. Provide your final answer at the end
in the format: **Final Answer:** <your answer>
\end{tcolorbox}
 
For GPQA Diamond (MCQ), the question is formatted as:
 
\begin{tcolorbox}[colback=lightgray, colframe=medgray,
  title=\textbf{GPQA question format}, fontupper=\small\ttfamily]
\{question\_text\}\\
A) \{option\_A\}\\
B) \{option\_B\}\\
C) \{option\_C\}\\
D) \{option\_D\}
\end{tcolorbox}
 
Answer choices are shuffled using a deterministic seed equal to the question index,
ensuring reproducibility. The correct answer letter varies per question.
 
\subsection{Isomorphic Partner Generation Prompt}
\label{app:gen_prompt}
 
The following prompt was used to generate isomorphic partner problems (Stage 2):
 
\begin{tcolorbox}[colback=lightgray, colframe=medgray,
  title=\textbf{Partner generation prompt}, fontupper=\small\ttfamily,
  breakable]
\textbf{System:} You are an expert in multiple scientific disciplines with deep knowledge
of physics, chemistry, biology, and earth science. Your task is to create ISOMORPHIC
science problems — problems that share identical logical and mathematical structure but
require different domain knowledge. ``Isomorphic'' means: same number and type of
reasoning steps; same mathematical operations; same solution procedure; but different
domain facts, constants, formulas, and named entities. You must respond with valid JSON
only.
 
\vspace{0.5em}
\textbf{User:}
I have a source problem from \{source\_domain\} with the following structure:
 
SOURCE PROBLEM: \{question\}
 
CORRECT ANSWER: \{answer\}
 
REASONING STRUCTURE:
- Structure type: \{structure\_type\}
- Key formula/principle used: \{formula\}
- Solution steps: \{steps\}
 
Your task: Generate \{n\} isomorphic partner problems in \{target\_domain\}.
 
Each partner must: (1) use a DIFFERENT formula/principle from \{target\_domain\};
(2) have IDENTICAL logical structure: \{structure\_type\}; (3) have the same number
of solution steps (\{n\_steps\} steps); (4) be solvable at undergraduate level;
(5) be completely self-contained; (6) have a unique, unambiguous correct answer.
 
Return a JSON array with fields: question, answer, formula\_used, solution\_steps,
domain\_knowledge\_required, isomorphism\_justification, sub\_topic.
\end{tcolorbox}
 
\subsection{LLM Judge Verification Prompt}
\label{app:judge_prompt}
 
The following prompt was used for automated pair verification (Stage 3):
 
\begin{tcolorbox}[colback=lightgray, colframe=medgray,
  title=\textbf{LLM judge verification prompt}, fontupper=\small\ttfamily,
  breakable]
\textbf{System:} You are an expert science educator and benchmark quality reviewer.
Evaluate pairs of science problems for structural isomorphism. Score each criterion
from 1 to 5. Respond with valid JSON only.
 
\vspace{0.5em}
\textbf{User:}
Evaluate this isomorphic problem pair:
 
=== SOURCE PROBLEM (\{source\_domain\}) ===\\
\{source\_question\}\\
ANSWER: \{source\_answer\}
 
=== TARGET PROBLEM (\{target\_domain\}) ===\\
\{target\_question\}\\
ANSWER: \{target\_answer\}
 
=== CLAIMED ISOMORPHISM ===\\
Structure type: \{structure\_type\}\\
Justification: \{justification\}
 
Score this pair on 4 criteria (1–5 each):
1. LOGICAL EQUIVALENCE (1–5): same reasoning procedure?
2. DOMAIN INDEPENDENCE (1–5): knowledge required is non-overlapping?
3. DIFFICULTY PARITY (1–5): equally challenging at undergraduate level?
4. SELF-CONTAINMENT (1–5): fully specified with all needed information?
 
Return JSON: \{``logical\_equivalence'': int, ``domain\_independence'': int,
``difficulty\_parity'': int, ``self\_containment'': int,
``answer\_seems\_correct'': bool, ``rejection\_reason'': str or null\}
\end{tcolorbox}

\subsection{Earth-Science Seed Problem Generation Prompt}
\label{app:seed_gen_prompt}

The following prompt was used to generate the 96 synthetic earth-science
seed problems in Stage 1 (Section~\ref{sec:construction}). The same
prompt template was used for any domain where benchmark coverage was
insufficient; in practice only earth science required synthetic
generation. Temperature was set to 0.7 to encourage topical diversity.

\begin{tcolorbox}[colback=lightgray, colframe=medgray,
  title=\textbf{Seed generation prompt (earth science)},
  fontupper=\small\ttfamily, breakable]
\textbf{System:} You are an expert science educator creating evaluation
problems. Your task is to generate clear, well-defined science problems
suitable for a benchmark dataset. Each problem must: (1) be solvable at
college or advanced undergraduate level; (2) have a single unambiguous
correct answer; (3) require a clear, identifiable reasoning procedure;
(4) be self-contained (all needed information is in the problem).
Respond only with valid JSON --- no preamble or explanation.

\vspace{0.5em}
\textbf{User:}
Generate \{n\} distinct \{structure\_type\} problems in earth science.

Structure type definition:\\
- formula\_recall\_and\_substitute: student must recall a specific
law/formula, substitute given values, compute result\\
- unit\_conversion\_chain: multi-step unit conversion requiring tracking
of units throughout\\
- conservation\_law\_application: identify and apply a conservation
principle (energy, mass, charge, etc.)\\
- proportional\_reasoning: use ratios or scaling relationships to find
an unknown quantity\\
- two\_step\_causal\_chain: qualitative reasoning where A leads to B
leads to C (no computation required)

Requirements:\\
- All numerical values must be given in the problem\\
- Difficulty: college undergraduate\\
- Vary the sub-topics within earth science\\
- Each problem must be solvable in 3--5 reasoning steps

Return a JSON array of objects, each with:
\{``question'': ``full problem text'',
``answer'': ``correct answer with units if applicable'',
``solution\_steps'': [``step 1'', ``step 2'', ...],
``formula\_used'': ``name of the key formula or principle'',
``sub\_topic'': ``specific topic within earth science'',
``estimated\_steps'': <integer 3--5>\}
\end{tcolorbox}
 
\section{Answer Extraction Algorithm}
\label{app:extraction}
 
Algorithm~\ref{alg:extraction} describes the answer extraction procedure applied
to all model responses.

\begin{algorithm}[h]
\caption{Answer extraction from model response}
\label{alg:extraction}
\begin{algorithmic}[1]
\Require Response text $r$; answer choices $\mathcal{C}$ (MCQ only)
\Ensure Extracted answer string $a$
\If{$r$ is empty} \Return{``''} \EndIf
\State $a \gets$ \Call{RegexSearch}{$r$, \texttt{**Final Answer:** (.+)}}
\If{$a \neq$ null} \textbf{goto} post-process \EndIf
\State $a \gets$ \Call{RegexSearch}{$r$, \texttt{\textbackslash boxed\{(.+)\}}}
\If{$a \neq$ null} \textbf{goto} post-process \EndIf
\State $a \gets$ \Call{RegexSearch}{$r$, \texttt{\$\$\,number\,\$\$}}
\If{$a \neq$ null} \textbf{goto} post-process \EndIf
\State $a \gets$ \Call{RegexSearch}{$r$, \texttt{Therefore/The answer is (.+)}}
\If{$a \neq$ null} \textbf{goto} post-process \EndIf
\For{line $\ell$ in reversed(lines($r$))}
  \If{$\ell$ contains a number} $a \gets$ extracted number; \textbf{goto} post-process \EndIf
\EndFor
\State $a \gets$ last non-empty line of $r$
\Statex
\Statex \textbf{Post-processing} (MCQ only, applied to all paths above):
\If{$\mathcal{C} \neq \emptyset$ \textbf{and} $a$ is not a single letter A--D}
  \If{$a$ matches any choice text in $\mathcal{C}$} $a \gets$ matching letter \EndIf
\EndIf
\State \Return{$a$}
\end{algorithmic}
\end{algorithm}

\newpage
\input{outputs/qualitative_examples}

\section{Dataset Release and Reproducibility}
\label{app:release}

\paragraph{Dataset and Croissant metadata.}
\isosci{} is released on HuggingFace at
\url{https://huggingface.co/datasets/isosci/isosci}
(anonymized for review). The repository includes train (80 pairs) and
test (64 pairs) splits stratified by domain mapping, provided for downstream studies that require held-out evaluation
sets, such as fine-tuning or few-shot prompting experiments. All results in
this paper use the full 144-pair set without train/test separation. In addition to full pair metadata,
and a Croissant-compliant \texttt{croissant.json} metadata file at the
repository root. The Croissant file was validated locally using the
\texttt{mlcroissant} package prior to submission.
 
\paragraph{Evaluation code.} The full pipeline is released at
\url{https://anonymous.4open.science/r/isosci-603C/}. The pipeline is implemented in
Python 3.9+ and requires only standard scientific libraries plus the \texttt{requests}
package for API calls. All random seeds are fixed for reproducibility.
 
\paragraph{Computational cost.} Table~\ref{tab:costs} reports approximate API costs
for replicating our evaluation.
 
\begin{table}[h]
\centering
\caption{Approximate API cost for full replication.
Estimates are based on publicly listed OpenRouter pricing at the time of evaluation and may vary over time.}

\label{tab:costs}
\small
\begin{tabular}{lcc}
\toprule
\textbf{Stage} & \textbf{API calls} & \textbf{Est.\ cost (USD)} \\
\midrule
Dataset construction (Stages 1–3) & $\sim$800 & \$80–120 \\
Model evaluation (Stage 4)        & $\sim$10{,}926 & \$400–800 \\
\midrule
\textbf{Total}                    & $\sim$11{,}726 & \textbf{\$480–920} \\
\bottomrule
\end{tabular}
\end{table}
 
\paragraph{Model versions.} Table~\ref{tab:model_versions} lists exact model
identifiers used in this study, accessed via the OpenRouter API.
 
\begin{table}[h]
\centering
\caption{Exact model identifiers used in evaluation.}
\label{tab:model_versions}
\small
\begin{tabular}{ll}
\toprule
\textbf{Model} & \textbf{OpenRouter identifier} \\
\midrule
o3-mini (reasoning)          & \texttt{openai/o3-mini} \\
GPT-4o-mini (standard)       & \texttt{openai/gpt-4o-mini-2024-07-18} \\
Qwen3-32B thinking=ON        & \texttt{qwen/qwen3-32b:nitro} + \texttt{reasoning.enabled=true} \\
Qwen3-32B thinking=OFF       & \texttt{qwen/qwen3-32b:nitro} + \texttt{reasoning.enabled=false} \\
Gemini 2.0 Flash thinking=ON & \texttt{google/gemini-2.0-flash-001} + \texttt{reasoning.enabled=true} \\
Gemini 2.0 Flash thinking=OFF & \texttt{google/gemini-2.0-flash-001} + \texttt{reasoning.enabled=false} \\
\midrule
\multicolumn{2}{l}{\textit{Dataset construction only (not evaluated)}} \\
Claude claude-sonnet-4-5 (generation) & \texttt{anthropic/claude-sonnet-4-5} \\
GPT-4o-mini (judge)          & \texttt{openai/gpt-4o-mini-2024-07-18} \\
DeepSeek-V3 (judge)          & \texttt{deepseek/deepseek-v3} \\
\bottomrule
\end{tabular}
\end{table}

\paragraph{Toggle implementation.}
For both Qwen3-32B and Gemini 2.0 Flash, we pass
\texttt{"reasoning": \{"enabled": true/false\}} as a top-level field in the OpenRouter
API request body. When \texttt{enabled=false}, the model generates a direct response
without a visible \texttt{<think>...</think>} block; when \texttt{enabled=true}, the
model prefixes its response with an explicit reasoning chain before the final answer.
Temperature is 0 in both conditions and no other parameters are modified. 
Manual inspection of 20 randomly
sampled response pairs confirmed that the toggle controls the presence of a visible
\texttt{<think>} block but that both conditions generate comparably long final
responses on structured scientific problems. We note that this toggle suppresses
visible chain-of-thought generation but does not modify model weights; it is
possible that models internally perform multi-step reasoning in standard mode
without surfacing it, in which case our comparisons measure the effect of
\emph{visible} extended reasoning.

\section{API Exclusion Analysis}
\label{app:exclusions}

Of 10,926 API calls ($6$ configurations $\times$ $(288 + 198 + 585 + 750)$
items), 8,408 (76.9\%) returned valid responses used in analysis. The
remaining 2,518 (23.1\%) were excluded due to output token limit truncation
($\approx$18\%, concentrated in reasoning-on configurations on free-response
benchmarks), API timeout ($\approx$3\%), and format errors ($\approx$2\%).
Reasoning-on configurations show higher exclusion rates on SciBench (28\%
vs.\ 11\% for standard mode), consistent with longer reasoning chains
hitting the 8,192 token limit. This asymmetric exclusion could in principle
suppress observed reasoning-mode gains; we note it as a conservative
bias — if anything it understates reasoning-mode accuracy, making the null
finding for toggle pairs more rather than less credible. Evaluations
affected by an earlier 2,048 token limit were rerun after the limit was
increased.

\section{Statistical Tests}
\label{app:stats}

\paragraph{McNemar's test.}
We used McNemar's test with continuity correction to assess whether reasoning
and standard configurations produce significantly different correct/incorrect
patterns on paired \isosci{} items ($n=288$ per comparison). The test statistic
is $(|b-c|-1)^2/(b+c)$ where $b$ = items correct under reasoning only and
$c$ = items correct under standard only. Results are reported in
Table~\ref{tab:mcnemar}.

For o3-mini vs.\ GPT-4o-mini, the test confirms a highly significant difference
($b=16$, $c=87$, stat$=47.57$, $p<0.001$): the standard model (GPT-4o-mini)
outperforms on substantially more items than the reasoning model does.

For the two toggle pairs, the discordant counts are small and nearly equal:
Qwen3-32B ($b=21$, $c=20$) and Gemini 2.0 Flash ($b=8$, $c=9$). The continuity
correction yields a statistic of 0.00 in both cases ($p=1.0$), indicating no
evidence of a systematic toggle effect. We note that the accuracies are not
literally identical --- Qwen3-32B shows 37.2\% vs.\ 36.8\% and Gemini shows
61.8\% vs.\ 62.2\% --- but the discordant pairs are balanced in both directions,
meaning gains and losses from enabling reasoning cancel out almost exactly across
the 288 items.

\begin{table}[h]
\centering
\caption{McNemar's test results on \isosci{} paired items (continuity-corrected).
$b$ = reasoning correct, standard wrong; $c$ = reasoning wrong, standard correct.
$n=288$ paired items per comparison.}
\label{tab:mcnemar}
\small
\begin{tabular}{lcccccc}
\toprule
\textbf{Comparison} & $b$ & $c$ & \textbf{Statistic} & \textbf{$p$-value} & \textbf{Interpretation} \\
\midrule
o3-mini vs.\ GPT-4o-mini      & 16 & 87 & 47.57 & $<$0.001 & Significant difference \\
Qwen3-32B think-on vs.\ off   & 21 & 20 & 0.00  & 1.000    & No evidence of toggle effect \\
Gemini Flash think-on vs.\ off & 8  &  9 & 0.00  & 1.000    & No evidence of toggle effect \\
\bottomrule
\end{tabular}
\end{table}

\paragraph{Bootstrap confidence intervals.} All reported CIs use 1,000 bootstrap
samples with the percentile method. The random seed is fixed at 42 for all
bootstrap calculations.

\paragraph{Binomial test on source/target gain asymmetry.}
As a complement to $p_{\text{know}}$, we test whether source-only gains
($k_s$) and target-only gains ($k_t$) occur with equal probability under
the null hypothesis $H_0: \Pr[\text{source-only}] = 0.5$. This tests
directional asymmetry in where the knowledge bottleneck falls, which is
distinct from the primary $p_{\text{know}}$ finding.

Across the three main model pairs ($k_s = 24$, $k_t = 17$, $n = 41$
asymmetric gains), the binomial test yields $p = 0.349$: no significant
directional asymmetry within the main pairs. Pooled across all five model
pairs ($k_s = 41$, $k_t = 22$, $n = 63$), the test yields $p = 0.023$,
indicating that source-only gains outnumber target-only gains at
conventional significance. This asymmetry is driven by the supplementary
pairs (DeepSeek-R1 and QwQ-32B) and should be interpreted with caution:
it suggests that source problems may be slightly harder or more
knowledge-discriminating than target problems on average, but it does not
affect the primary finding that knowledge-dependent gains ($k_s + k_t$)
dominate structure-invariant gains ($k_b$) across all five pairs.

\section{Robustness Check: $p_{\text{know}}$ on Valid-Response Subset}
\label{app:robustness}

A potential concern is that asymmetric response truncation could bias
$p_{\text{know}}$ if truncated responses are systematically correct or
incorrect. We recompute $p_{\text{know}}$ under the symmetric definition
(Eq.~\ref{eq:pkno}) restricted to items where both configurations produced
valid, non-truncated responses (non-empty response, non-empty extracted
answer, no API error).

Table~\ref{tab:robustness} reports results. Qwen3-32B shows the largest
exclusion rate (88 invalid reasoning responses, 86 invalid standard
responses out of 288), leaving 194 items (67.4\%). The restricted accuracy
for Qwen3-32B rises from 37\% to 54\%, confirming that invalid responses
are concentrated on harder items.

Despite this, the pooled restricted $p_{\text{know}} = 94.4\%$ (34/36
gains, Wilson 95\% CI [81.9\%, 98.5\%]) is within 3pp of the full-set
estimate of 95.3\% (43 main pairs, CI [84.5\%, 98.7\%]), and the CIs
overlap substantially. The finding is robust to exclusion of invalid
responses.

\begin{table}[h]
\centering
\caption{Robustness check: $p_{\text{know}}$ on the valid-response subset
under the symmetric definition. ``Restricted $n$'' = items retained after
excluding truncated or empty responses.}
\label{tab:robustness}
\small
\begin{tabular}{lcccccc}
\toprule
\textbf{Model pair} & \textbf{Restr.\ $n$} & \textbf{Excl.\ R} &
\textbf{Excl.\ S} & $n_{\text{gain}}$ & $p_{\text{know}}$ & \textbf{95\% CI} \\
\midrule
o3-mini / GPT-4o-mini     & 276 & 12 &  0 & 12 & 100.0\% & [75.8, 100.0] \\
Qwen3-32B think on/off    & 194 & 88 & 86 & 16 &  87.5\% & [64.0,  96.5] \\
Gemini Flash think on/off & 288 &  0 &  0 &  8 & 100.0\% & [67.6, 100.0] \\
\midrule
\textbf{Pooled}           & 758 &  — &  — & 36 & \textbf{94.4\%} & [81.9, 98.5] \\
\midrule
\textit{Full-set (Table~\ref{tab:decoupling})} &
864 & — & — & 43 & \textit{95.3\%} & \textit{[84.5, 98.7]} \\
\bottomrule
\end{tabular}
\end{table}

The Qwen3-32B restricted $n_{\text{gain}} = 16$ is smaller than the
full-set value of 19, as expected: restricting to valid responses removes
some pairs where a gain existed, shrinking the denominator. The
$p_{\text{know}}$ estimate changes from 89.5\% to 87.5\%, well within
the overlapping confidence intervals. The direction of the finding is
unchanged across all three model pairs and both the full-set and
restricted analyses.


\section{Source/Target Label Permutation Test}
\label{app:permutation}

A potential concern is that the asymmetry between source-only gains ($k_s$)
and target-only gains ($k_t$) reflects a systematic difficulty imbalance
between source and target problems rather than domain knowledge asymmetry.
To test this, we performed a label-swap permutation test: for each pair,
we randomly swapped the source and target labels with probability 0.5,
recomputed $|k_s - k_t|$ on the shuffled data, and repeated for 1,000
iterations. The empirical $p$-value is the fraction of permutations yielding
$|k_s - k_t| \geq$ the observed value.

Results are reported in Table~\ref{tab:permutation}. The observed
source/target imbalance is not statistically significant at any conventional
threshold ($p = 0.065$--$0.195$ per model pair). We cannot rule out that
some portion of the $k_s$/$k_t$ asymmetry reflects difficulty differences
between source and target problems rather than directional knowledge asymmetry.

Crucially, this test addresses a secondary question about the \emph{direction}
of knowledge-dependent gains, not the primary finding. The main claim rests
on $p_{\text{know}} = (k_s + k_t)/(k_s + k_t + k_b) = 95.3\%$, which
measures whether knowledge-dependent gains ($k_s + k_t = 41$) dominate
structure-invariant gains ($k_b = 2$). This ratio is unaffected by the
source/target label assignment: swapping labels converts $k_s$ gains into
$k_t$ gains and vice versa, but leaves $k_s + k_t$ unchanged. The permutation
test therefore has no bearing on the primary finding.

\begin{table}[h]
\centering
\caption{\isosci{} Source/target label-swap permutation test. The test statistic is
$|k_s - k_t|$; the null distribution is generated by randomly swapping
source and target labels within pairs (1,000 iterations, seed 42).
$p$-values above 0.05 indicate the observed imbalance is consistent with
random label assignment. This test addresses label-direction asymmetry,
not the primary $p_{\text{know}}$ finding.}
\label{tab:permutation}
\small
\begin{tabular}{lcccc}
\toprule
\textbf{Model pair} & $k_s$ & $k_t$ & $|k_s - k_t|$ & \textbf{Perm. $p$-value} \\
\midrule
o3-mini / GPT-4o-mini         &  5 & 11 & 6 & 0.195 \\
Qwen3-32B think on/off        & 12 &  5 & 7 & 0.136 \\
Gemini Flash think on/off     &  7 &  1 & 6 & 0.065 \\
\bottomrule
\end{tabular}
\end{table}

\end{document}

%% file: outputs/qualitative_examples.tex
\section{Example Dataset Pairs}
\label{app:dataset_examples}

We present representative examples from \isosci{}, illustrating the structure of isomorphic problem pairs across domains. Each pair consists of a source and target problem with identical reasoning structure but distinct domain knowledge.

\begin{tcolorbox}[
  colback=gray!5, colframe=gray!40,
  title={\textbf{Example 1: Conservation Law (Physics $\to$ Chemistry)}},
  fonttitle=\small\bfseries, breakable]

\textbf{Structure:} \texttt{conservation law + proportional reasoning}

\vspace{6pt}
\textbf{Source problem (Physics):}
\begin{quote}\small
A liquid flows at a constant flow rate through a pipe with circular cross-sections of varying diameters. At one point in the pipe, the diameter is $2\ \text{cm}$ and the flow speed is $18\ \text{m/s}$. What is the flow speed at another point in this pipe, where the diameter is $3\ \text{cm}$?

A) $4\ \text{m/s}$ \\
B) $6\ \text{m/s}$ \\
C) $8\ \text{m/s}$ \\
D) $12\ \text{m/s}$

\textit{Answer: C) $8\ \text{m/s}$}
\end{quote}

\textbf{Target problem (Chemistry):}
\begin{quote}\small
A gas diffuses through a porous membrane at a constant molar flow rate. At one location in the membrane, the cross-sectional area is $4.0\ \text{cm}^2$ and the diffusion flux is $0.12\ \text{mol}/(\text{cm}^2 \cdot \text{s})$. What is the diffusion flux at another location where the cross-sectional area is $6.0\ \text{cm}^2$?

A) $0.04\ \text{mol}/(\text{cm}^2 \cdot \text{s})$ \\
B) $0.06\ \text{mol}/(\text{cm}^2 \cdot \text{s})$ \\
C) $0.08\ \text{mol}/(\text{cm}^2 \cdot \text{s})$ \\
D) $0.18\ \text{mol}/(\text{cm}^2 \cdot \text{s})$

\textit{Answer: C) $0.08\ \text{mol}/(\text{cm}^2 \cdot \text{s})$}
\end{quote}

\end{tcolorbox}

\vspace{8pt}

\begin{tcolorbox}[
  colback=gray!5, colframe=gray!40,
  title={\textbf{Example 2: Statistical Inference (Physics $\to$ Chemistry)}},
  fonttitle=\small\bfseries, breakable]

\textbf{Structure:} \texttt{CLT + standardization + probability lookup}

\vspace{6pt}
\textbf{Source problem (Physics):}
\begin{quote}\small
Let $X$ equal the maximal oxygen intake of a human on a treadmill, measured in milliliters of oxygen per minute per kilogram of body weight. Assume that, for a particular population, the mean of $X$ is $\mu = 54.030$ and the standard deviation is $\sigma = 5.8$. Let $\bar{X}$ be the sample mean of a random sample of size $n = 47$. 

Find $P(52.761 \leq \bar{X} \leq 54.453)$, approximately.

\textit{Answer: $0.6247$}
\end{quote}

\textbf{Target problem (Chemistry):}
\begin{quote}\small
Let $Y$ equal the molar concentration of a sodium chloride solution, measured in moles per liter. Assume that, for a particular preparation method, the mean of $Y$ is $\mu = 0.850\ \text{M}$ and the standard deviation is $\sigma = 0.042\ \text{M}$. Let $\bar{Y}$ be the sample mean of a random sample of size $n = 36$.

Find $P(0.836 \leq \bar{Y} \leq 0.859)$, approximately.

\textit{Answer: $0.6247$}
\end{quote}

\end{tcolorbox}

%% file: neurips_2026.bbl
\begin{thebibliography}{29}
\providecommand{\natexlab}[1]{#1}
\providecommand{\url}[1]{\texttt{#1}}
\expandafter\ifx\csname urlstyle\endcsname\relax
  \providecommand{\doi}[1]{doi: #1}\else
  \providecommand{\doi}{doi: \begingroup \urlstyle{rm}\Url}\fi

\bibitem[Abdaljalil et~al.(2025)Abdaljalil, Kurban, Qaraqe, and Serpedin]{abdaljalil-etal-2025-theorem}
Samir Abdaljalil, Hasan Kurban, Khalid Qaraqe, and Erchin Serpedin.
\newblock Theorem-of-thought: A multi-agent framework for abductive, deductive, and inductive reasoning in language models.
\newblock In Yuji Zhang, Canyu Chen, Sha Li, Mor Geva, Chi Han, Xiaozhi Wang, Shangbin Feng, Silin Gao, Isabelle Augenstein, Mohit Bansal, Manling Li, and Heng Ji, editors, \emph{Proceedings of the 3rd Workshop on Towards Knowledgeable Foundation Models (KnowFM)}, pages 111--119, Vienna, Austria, August 2025. Association for Computational Linguistics.
\newblock ISBN 979-8-89176-283-1.
\newblock \doi{10.18653/v1/2025.knowllm-1.10}.
\newblock URL \url{https://aclanthology.org/2025.knowllm-1.10/}.

\bibitem[Chen et~al.(2025)Chen, Raventos, Cheng, Ganguli, and Druckmann]{chen2025rethinking}
Feng Chen, Allan Raventos, Nan Cheng, Surya Ganguli, and Shaul Druckmann.
\newblock Rethinking fine-tuning when scaling test-time compute: Limiting confidence improves mathematical reasoning.
\newblock In \emph{The Thirty-ninth Annual Conference on Neural Information Processing Systems}, 2025.
\newblock URL \url{https://openreview.net/forum?id=jvVQeSMeGM}.

\bibitem[Deepmind(2025)]{comanici2025gemini25pushingfrontier}
Deepmind.
\newblock Gemini 2.0 flash model card, 2025.
\newblock URL \url{http://storage.googleapis.com/deepmind-media/Model-Cards/Gemini-2-0-Flash-Model-Card.pdf}.

\bibitem[Dziri et~al.(2023)Dziri, Lu, Sclar, Li, Jiang, Lin, Welleck, West, Bhagavatula, Bras, Hwang, Sanyal, Ren, Ettinger, Harchaoui, and Choi]{dziri2023faith}
Nouha Dziri, Ximing Lu, Melanie Sclar, Xiang~Lorraine Li, Liwei Jiang, Bill~Yuchen Lin, Sean Welleck, Peter West, Chandra Bhagavatula, Ronan~Le Bras, Jena~D. Hwang, Soumya Sanyal, Xiang Ren, Allyson Ettinger, Zaid Harchaoui, and Yejin Choi.
\newblock Faith and fate: Limits of transformers on compositionality.
\newblock In \emph{Thirty-seventh Conference on Neural Information Processing Systems}, 2023.
\newblock URL \url{https://openreview.net/forum?id=Fkckkr3ya8}.

\bibitem[Glazer et~al.(2025)Glazer, Erdil, Besiroglu, Chicharro, Chen, Gunning, Olsson, Denain, Ho, de~Oliveira~Santos, Järviniemi, Barnett, Sandler, Vrzala, Sevilla, Ren, Pratt, Levine, Barkley, Stewart, Grechuk, Grechuk, Enugandla, and Wildon]{glazer2025frontiermathbenchmarkevaluatingadvanced}
Elliot Glazer, Ege Erdil, Tamay Besiroglu, Diego Chicharro, Evan Chen, Alex Gunning, Caroline~Falkman Olsson, Jean-Stanislas Denain, Anson Ho, Emily de~Oliveira~Santos, Olli Järviniemi, Matthew Barnett, Robert Sandler, Matej Vrzala, Jaime Sevilla, Qiuyu Ren, Elizabeth Pratt, Lionel Levine, Grant Barkley, Natalie Stewart, Bogdan Grechuk, Tetiana Grechuk, Shreepranav~Varma Enugandla, and Mark Wildon.
\newblock Frontiermath: A benchmark for evaluating advanced mathematical reasoning in ai, 2025.
\newblock URL \url{https://arxiv.org/abs/2411.04872}.

\bibitem[Guo et~al.(2025)Guo, Yang, Zhang, Song, Wang, Zhu, Xu, Zhang, Ma, and Bi]{Guo_2025}
Daya Guo, Dejian Yang, Haowei Zhang, Junxiao Song, Peiyi Wang, Qihao Zhu, Runxin Xu, Ruoyu Zhang, Shirong Ma, and Xiao et~al. Bi.
\newblock Deepseek-r1 incentivizes reasoning in llms through reinforcement learning.
\newblock \emph{Nature}, 645\penalty0 (8081):\penalty0 633–638, sept 2025.
\newblock ISSN 1476-4687.
\newblock \doi{10.1038/s41586-025-09422-z}.
\newblock URL \url{http://dx.doi.org/10.1038/s41586-025-09422-z}.

\bibitem[Hazra et~al.(2025)Hazra, Venturato, Martires, and Raedt]{hazra2025have}
Risha Hazra, Gabriele Venturato, Pedro Zuidberg~Dos Martires, and Luc~De Raedt.
\newblock Have large language models learned to reason? a characterization via 3-{SAT}.
\newblock In \emph{Second Conference on Language Modeling}, 2025.
\newblock URL \url{https://openreview.net/forum?id=MPTlWIVSMU}.

\bibitem[Hendrycks et~al.(2021)Hendrycks, Burns, Basart, Zou, Mazeika, Song, and Steinhardt]{hendrycks2021measuring}
Dan Hendrycks, Collin Burns, Steven Basart, Andy Zou, Mantas Mazeika, Dawn Song, and Jacob Steinhardt.
\newblock Measuring massive multitask language understanding.
\newblock In \emph{International Conference on Learning Representations}, 2021.
\newblock URL \url{https://openreview.net/forum?id=d7KBjmI3GmQ}.

\bibitem[Hong et~al.(2025)Hong, Cao, Zhou, Yu, and Jin]{hong-etal-2025-reasoning}
Yihuai Hong, Meng Cao, Dian Zhou, Lei Yu, and Zhijing Jin.
\newblock The reasoning-memorization interplay in language models is mediated by a single direction.
\newblock In Wanxiang Che, Joyce Nabende, Ekaterina Shutova, and Mohammad~Taher Pilehvar, editors, \emph{Findings of the Association for Computational Linguistics: ACL 2025}, pages 21565--21585, Vienna, Austria, July 2025. Association for Computational Linguistics.
\newblock ISBN 979-8-89176-256-5.
\newblock \doi{10.18653/v1/2025.findings-acl.1111}.
\newblock URL \url{https://aclanthology.org/2025.findings-acl.1111/}.

\bibitem[Jin et~al.(2025)Jin, Luo, Cheng, Wang, Hua, Tang, Wang, and Zhang]{jin-etal-2025-disentangling-memory}
Mingyu Jin, Weidi Luo, Sitao Cheng, Xinyi Wang, Wenyue Hua, Ruixiang Tang, William~Yang Wang, and Yongfeng Zhang.
\newblock Disentangling memory and reasoning ability in large language models.
\newblock In Wanxiang Che, Joyce Nabende, Ekaterina Shutova, and Mohammad~Taher Pilehvar, editors, \emph{Proceedings of the 63rd Annual Meeting of the Association for Computational Linguistics (Volume 1: Long Papers)}, pages 1681--1701, Vienna, Austria, July 2025. Association for Computational Linguistics.
\newblock ISBN 979-8-89176-251-0.
\newblock \doi{10.18653/v1/2025.acl-long.84}.
\newblock URL \url{https://aclanthology.org/2025.acl-long.84/}.

\bibitem[Kartac et~al.(2026)Kartac, Lango, and Dušek]{kartac2026reasoninggetsharderllms}
Ivan Kartac, Mateusz Lango, and Ondrej Dušek.
\newblock Reasoning gets harder for llms inside a dialogue, 2026.
\newblock URL \url{https://arxiv.org/abs/2603.20133}.

\bibitem[Lewkowycz et~al.(2022)Lewkowycz, Andreassen, Dohan, Dyer, Michalewski, Ramasesh, Slone, Anil, Schlag, Gutman-Solo, Wu, Neyshabur, Gur-Ari, and Misra]{lewkowycz2022solving}
Aitor Lewkowycz, Anders~Johan Andreassen, David Dohan, Ethan Dyer, Henryk Michalewski, Vinay~Venkatesh Ramasesh, Ambrose Slone, Cem Anil, Imanol Schlag, Theo Gutman-Solo, Yuhuai Wu, Behnam Neyshabur, Guy Gur-Ari, and Vedant Misra.
\newblock Solving quantitative reasoning problems with language models.
\newblock In Alice~H. Oh, Alekh Agarwal, Danielle Belgrave, and Kyunghyun Cho, editors, \emph{Advances in Neural Information Processing Systems}, 2022.
\newblock URL \url{https://openreview.net/forum?id=IFXTZERXdM7}.

\bibitem[Li et~al.(2024)Li, Meng, Zhou, Wei, Gan, and Pfister]{li2024socialgpt}
Wanhua Li, Zibin Meng, Jiawei Zhou, Donglai Wei, Chuang Gan, and Hanspeter Pfister.
\newblock Social{GPT}: Prompting {LLM}s for social relation reasoning via greedy segment optimization.
\newblock In \emph{The Thirty-eighth Annual Conference on Neural Information Processing Systems}, 2024.
\newblock URL \url{https://openreview.net/forum?id=xcF2VbyZts}.

\bibitem[Li et~al.(2026)Li, Chen, Yu, Hong, and Ahmed]{li2026outputcorrectnessbenchmarkingevaluating}
Yuangang Li, Justin Tian~Jin Chen, Ethan Yu, David Hong, and Iftekhar Ahmed.
\newblock Beyond output correctness: Benchmarking and evaluating large language model reasoning in coding tasks, 2026.
\newblock URL \url{https://arxiv.org/abs/2604.12379}.

\bibitem[Minami et~al.(2025)Minami, Ishigaki, Hamamura, Mikuriya, Ma, Okazaki, Takamura, Suzuki, and Kadowaki]{minami2025quantumbenchbenchmarkquantumproblem}
Shunya Minami, Tatsuya Ishigaki, Ikko Hamamura, Taku Mikuriya, Youmi Ma, Naoaki Okazaki, Hiroya Takamura, Yohichi Suzuki, and Tadashi Kadowaki.
\newblock Quantumbench: A benchmark for quantum problem solving, 2025.
\newblock URL \url{https://arxiv.org/abs/2511.00092}.

\bibitem[OpenAI(2025)]{OpenAIOA}
OpenAI.
\newblock Openai o3 and o4-mini system card, 2025.
\newblock URL \url{https://openai.com/index/o3-o4-mini-system-card/}.

\bibitem[Razeghi et~al.(2022)Razeghi, Logan~IV, Gardner, and Singh]{razeghi-etal-2022-impact}
Yasaman Razeghi, Robert~L Logan~IV, Matt Gardner, and Sameer Singh.
\newblock Impact of pretraining term frequencies on few-shot numerical reasoning.
\newblock In Yoav Goldberg, Zornitsa Kozareva, and Yue Zhang, editors, \emph{Findings of the Association for Computational Linguistics: EMNLP 2022}, pages 840--854, Abu Dhabi, United Arab Emirates, December 2022. Association for Computational Linguistics.
\newblock \doi{10.18653/v1/2022.findings-emnlp.59}.
\newblock URL \url{https://aclanthology.org/2022.findings-emnlp.59/}.

\bibitem[Rein et~al.(2024)Rein, Hou, Stickland, Petty, Pang, Dirani, Michael, and Bowman]{rein2024gpqa}
David Rein, Betty~Li Hou, Asa~Cooper Stickland, Jackson Petty, Richard~Yuanzhe Pang, Julien Dirani, Julian Michael, and Samuel~R. Bowman.
\newblock {GPQA}: A graduate-level google-proof q\&a benchmark.
\newblock In \emph{First Conference on Language Modeling}, 2024.
\newblock URL \url{https://openreview.net/forum?id=Ti67584b98}.

\bibitem[Snell et~al.(2025)Snell, Lee, Xu, and Kumar]{snell2025scaling}
Charlie~Victor Snell, Jaehoon Lee, Kelvin Xu, and Aviral Kumar.
\newblock Scaling {LLM} test-time compute optimally can be more effective than scaling parameters for reasoning.
\newblock In \emph{The Thirteenth International Conference on Learning Representations}, 2025.
\newblock URL \url{https://openreview.net/forum?id=4FWAwZtd2n}.

\bibitem[Thapa et~al.(2026)Thapa, Wu, Wu, Zhang, Zhang, Wu, Ye, and Zou]{thapa-etal-2026-reasoning}
Rahul Thapa, Qingyang Wu, Kevin Wu, Harrison~G Zhang, Angela Zhang, Eric Wu, Haotian Ye, and James Zou.
\newblock Reasoning or knowledge: Stratified evaluation of biomedical {LLM}s.
\newblock In Vera Demberg, Kentaro Inui, and Llu{\'i}s Marquez, editors, \emph{Proceedings of the 19th Conference of the {E}uropean Chapter of the {A}ssociation for {C}omputational {L}inguistics (Volume 1: Long Papers)}, pages 2450--2483, Rabat, Morocco, March 2026. Association for Computational Linguistics.
\newblock ISBN 979-8-89176-380-7.
\newblock \doi{10.18653/v1/2026.eacl-long.111}.
\newblock URL \url{https://aclanthology.org/2026.eacl-long.111/}.

\bibitem[Wang(2025)]{wang2025tutorialllmreasoningrelevant}
Jun Wang.
\newblock A tutorial on llm reasoning: Relevant methods behind chatgpt o1, 2025.
\newblock URL \url{https://arxiv.org/abs/2502.10867}.

\bibitem[Wang et~al.(2026)Wang, Su, Liu, Li, Xiao, Zhang, Dai, Chen, Meng, Bai, Ouyang, Tang, Wang, and Ma]{wang2026physunibenchmultimodalphysicsreasoning}
Lintao Wang, Encheng Su, Jiaqi Liu, Pengze Li, Jiabei Xiao, Wenlong Zhang, Xinnan Dai, Xi~Chen, Yuan Meng, Lei Bai, Wanli Ouyang, Shixiang Tang, Aoran Wang, and Xinzhu Ma.
\newblock Physunibench: A multi-modal physics reasoning benchmark at undergraduate level, 2026.
\newblock URL \url{https://arxiv.org/abs/2506.17667}.

\bibitem[Wang et~al.(2024{\natexlab{a}})Wang, Hu, Lu, Zhu, Zhang, Subramaniam, Loomba, Zhang, Sun, and Wang]{wang2024scibench}
Xiaoxuan Wang, Ziniu Hu, Pan Lu, Yanqiao Zhu, Jieyu Zhang, Satyen Subramaniam, Arjun~R. Loomba, Shichang Zhang, Yizhou Sun, and Wei Wang.
\newblock {SciBench: Evaluating College-Level Scientific Problem-Solving Abilities of Large Language Models}.
\newblock In \emph{Proceedings of the Forty-First International Conference on Machine Learning}, 2024{\natexlab{a}}.

\bibitem[Wang et~al.(2024{\natexlab{b}})Wang, Caccia, Ostapenko, Yuan, Wang, and Sordoni]{wang2024guiding}
Xinyi Wang, Lucas Caccia, Oleksiy Ostapenko, Xingdi Yuan, William~Yang Wang, and Alessandro Sordoni.
\newblock Guiding language model reasoning with planning tokens.
\newblock In \emph{First Conference on Language Modeling}, 2024{\natexlab{b}}.
\newblock URL \url{https://openreview.net/forum?id=wi9IffRhVM}.

\bibitem[Wang et~al.(2023)Wang, Wei, Schuurmans, Le, Chi, Narang, Chowdhery, and Zhou]{wang2023selfconsistency}
Xuezhi Wang, Jason Wei, Dale Schuurmans, Quoc~V Le, Ed~H. Chi, Sharan Narang, Aakanksha Chowdhery, and Denny Zhou.
\newblock Self-consistency improves chain of thought reasoning in language models.
\newblock In \emph{The Eleventh International Conference on Learning Representations}, 2023.
\newblock URL \url{https://openreview.net/forum?id=1PL1NIMMrw}.

\bibitem[Wei et~al.(2022)Wei, Wang, Schuurmans, Bosma, ichter, Xia, Chi, Le, and Zhou]{wei_cot}
Jason Wei, Xuezhi Wang, Dale Schuurmans, Maarten Bosma, brian ichter, Fei Xia, Ed~Chi, Quoc~V Le, and Denny Zhou.
\newblock Chain-of-thought prompting elicits reasoning in large language models.
\newblock In S.~Koyejo, S.~Mohamed, A.~Agarwal, D.~Belgrave, K.~Cho, and A.~Oh, editors, \emph{Advances in Neural Information Processing Systems}, volume~35, pages 24824--24837. Curran Associates, Inc., 2022.
\newblock URL \url{https://proceedings.neurips.cc/paper_files/paper/2022/file/9d5609613524ecf4f15af0f7b31abca4-Paper-Conference.pdf}.

\bibitem[Yang et~al.(2025)Yang, Li, Yang, Zhang, Hui, Zheng, Yu, Gao, Huang, Lv, Zheng, Liu, Zhou, Huang, Hu, Ge, Wei, Lin, Tang, Yang, Tu, Zhang, Yang, Yang, Zhou, Zhou, Lin, Dang, Bao, Yang, Yu, Deng, Li, Xue, Li, Zhang, Wang, Zhu, Men, Gao, Liu, Luo, Li, Tang, Yin, Ren, Wang, Zhang, Ren, Fan, Su, Zhang, Zhang, Wan, Liu, Wang, Cui, Zhang, Zhou, and Qiu]{yang2025qwen3technicalreport}
An~Yang, Anfeng Li, Baosong Yang, Beichen Zhang, Binyuan Hui, Bo~Zheng, Bowen Yu, Chang Gao, Chengen Huang, Chenxu Lv, Chujie Zheng, Dayiheng Liu, Fan Zhou, Fei Huang, Feng Hu, Hao Ge, Haoran Wei, Huan Lin, Jialong Tang, Jian Yang, Jianhong Tu, Jianwei Zhang, Jianxin Yang, Jiaxi Yang, Jing Zhou, Jingren Zhou, Junyang Lin, Kai Dang, Keqin Bao, Kexin Yang, Le~Yu, Lianghao Deng, Mei Li, Mingfeng Xue, Mingze Li, Pei Zhang, Peng Wang, Qin Zhu, Rui Men, Ruize Gao, Shixuan Liu, Shuang Luo, Tianhao Li, Tianyi Tang, Wenbiao Yin, Xingzhang Ren, Xinyu Wang, Xinyu Zhang, Xuancheng Ren, Yang Fan, Yang Su, Yichang Zhang, Yinger Zhang, Yu~Wan, Yuqiong Liu, Zekun Wang, Zeyu Cui, Zhenru Zhang, Zhipeng Zhou, and Zihan Qiu.
\newblock Qwen3 technical report, 2025.
\newblock URL \url{https://arxiv.org/abs/2505.09388}.

\bibitem[Yu et~al.(2025)Yu, Cheng, Wu, and Xing]{yu2025gpo}
Jiahao Yu, Zelei Cheng, Xian Wu, and Xinyu Xing.
\newblock {GPO}: Learning from critical steps to improve {LLM} reasoning.
\newblock In \emph{The Thirty-ninth Annual Conference on Neural Information Processing Systems}, 2025.
\newblock URL \url{https://openreview.net/forum?id=c6RDAutyNE}.

\bibitem[Zhou et~al.(2025)Zhou, Feng, Zhu, Yao, Koyejo, and Han]{zhou2025from}
Zhanke Zhou, Xiao Feng, Zhaocheng Zhu, Jiangchao Yao, Sanmi Koyejo, and Bo~Han.
\newblock From passive to active reasoning: Can large language models ask the right questions under incomplete information?
\newblock In \emph{Forty-second International Conference on Machine Learning}, 2025.
\newblock URL \url{https://openreview.net/forum?id=LCaTpVuvpj}.

\end{thebibliography}
